\pdfoutput=1

\documentclass[11pt, dvipsnames]{article}

\usepackage[]{acl}

\usepackage{times}
\usepackage{latexsym}
\usepackage{graphicx}
\usepackage{xcolor}
\usepackage{soul}
\usepackage{colortbl}
\usepackage{color}
\usepackage{tabularx}
\usepackage{dingbat}
\usepackage{pifont}
\usepackage{amsmath}
\usepackage{amssymb}
\usepackage{makecell}
\usepackage{arydshln}
\usepackage{stfloats}
\usepackage{subfig}

\usepackage[T1]{fontenc}
\usepackage{CJKutf8}
\usepackage{multirow}

\usepackage{array}   
\usepackage[utf8]{inputenc}

\usepackage{microtype}
\usepackage{booktabs}
\usepackage{amsmath}

\definecolor{myblue}{rgb}{0.82, 0.94, 0.75}

\title{Toward Human-Like Evaluation for Natural Language Generation with Error Analysis}

\author{%
  Qingyu~Lu$^{1,2}$\thanks{~~Work was done when Qingyu was interning at JD Explore Academy.},
  Liang~Ding$^{2}$,
  Liping Xie$^{1}$\thanks{~~Corresponding Author.},
  Kanjian Zhang$^{1}$,
  Derek F. Wong$^{3}$,
  Dacheng Tao$^{2}$\\
  $^{1}$School of Automation, Southeast University\\
  $^{2}$JD~Explore~Academy $^{3}$NLP$^2$CT Lab, University of Macau \\
  \texttt{\{luqingyu,lpxie,kjzhang\}@seu.edu.cn}, \texttt{dingliang1@jd.com}, \\
  \texttt{derekfw@um.edu.mo}, 
  \texttt{dacheng.tao@gmail.com} \\}

\begin{document}
\maketitle
\begin{abstract}
The state-of-the-art language model-based automatic metrics, e.g. BARTScore, benefiting from large-scale contextualized pre-training, have been successfully used in a wide range of natural language generation (NLG) tasks, including machine translation, text summarization, and data-to-text. Recent studies show that considering both major errors (e.g. mistranslated tokens) and minor errors (e.g. imperfections in fluency) can produce high-quality human judgments. This inspires us to approach the final goal of the evaluation metrics (human-like evaluations) by automatic error analysis. To this end, we augment BARTScore by incorporating the human-like error analysis strategies, namely BARTScore++, where the final score consists of both the evaluations of major errors and minor errors. Experimental results show that BARTScore++ can consistently improve the performance of vanilla BARTScore and outperform existing top-scoring metrics in 20 out of 25 test settings. We hope our technique can also be extended to other pre-trained model-based metrics. We will release our code and scripts to facilitate the community.

\end{abstract}

\section{Introduction}

Implementations of large pre-trained language models (PLMs) have been proven effective in evaluating natural language generation (NLG) tasks \citep{mathur2020results, ma-etal-2019-results}. Metrics like BERTScore \citep{zhang2020bertscore} and Moverscore \citep{zhao-etal-2019-moverscore} leverage contextual embeddings provided by PLMs to evaluate the semantic similarity of sentences. Regression-based metrics like COMET \citep{rei-etal-2020-comet} and BLEURT \citep{sellam-etal-2020-bleurt} introduce a regression layer following PLMs to learn a supervised prediction using human evaluation. Recently, another line of research focuses on generation probabilities of seq2seq PLMs (consisting of both encoder and decoder) to measure the quality of generated texts, such as PRISM \citep{thompson-post-2020-automatic} and BARTScore \citep{yuan2021bartscore}.

It is commonly agreed that the ultimate goal of automatic evaluation is to achieve consistency with humans, namely \textit{human-like evaluation}. However, most recent studies show that the quality of human judgments can be improved through error analysis, incorporated in an error-based framework Multidimensional Quality Metric (MQM) \citep{freitag-etal-2021-experts}. MQM requires evaluators to identify errors and categorize them into different levels according to their severity. For instance, mistranslation \citep{weng-etal-2020-towards} and hallucination \citep{zhou-etal-2021-detecting} are mostly considered as \textit{Major} errors, and imperfections in fluency \citep{chow-etal-2019-wmdo} are often marked as \textit{Minor} errors. Different weights are then assigned to Major/ Minor errors, resulting in high-quality human evaluation scoring.

Analogous to Major/ Minor errors in MQM, we incorporate the evaluation of Explicit/ Implicit Errors into BARTScore, a state-of-the-art metric for NLG by \citet{yuan2021bartscore} and propose a metric called BARTScore++. We present an overview of our proposed method in Figure~\ref{fig:bartscore++}. Since explicit errors can be easily identified, we design an error analysis framework to generate a refined sentence where explicit errors are corrected. This helps to measure the distance of explicit/ implicit errors in vanilla BARTScore. BARTScore++ finally integrates these two types of errors by assigning weights to them respectively.

We experiment on machine translation (MT), text summarization (SUM), and data-to-text (D2T) tasks, and show that BARTScore++ consistently improves the performance of vanilla BARTScore, and surpass existing top-scoring metrics in 20 out of 25 test settings, even exceeding human performance on summarization dataset \texttt{Rank19}. We give further analyses to confirm that the consistent improvements come from the human-like (specifically, MQM-like) error judgment.

Our \textbf{main contributions} are as follows:

\begin{itemize}
    \item To the best of our knowledge, we take the first step toward human-like evaluation by incorporating error analysis mechanisms into existing advanced automatic metrics, e.g. BARTScore.
    
    \item We propose BARTScore++ using a novel error analysis framework to refine sentences and consider both the influence of explicit and implicit errors.

    \item We validate the effectiveness and universality of our BARTScore++ on a broad range of NLG evaluation tasks, achieving the SOTA in 20 out of 25 settings.
\end{itemize}

\section{Preliminaries}

\subsection{Problem Formulation}

For evaluation tasks of NLG mentioned in this paper, the goal is to acquire a score measuring the quality of generated text $\boldsymbol{y}$ given a reference signal $\boldsymbol{r}$. Unless otherwise stated, $\boldsymbol{r}$ represents the sentence properly created by human experts to assist in evaluation, and $\boldsymbol{y} = \left(y_1y_2\dots y_N\right)$, called hypothesis in this paper, refers to the generated text to be evaluated\footnote{Note that in text summarization evaluations, BARTScore may use the source sentence as the reference signal.}.

\begin{figure*}[t]
    \centering
    \includegraphics[scale=0.42]{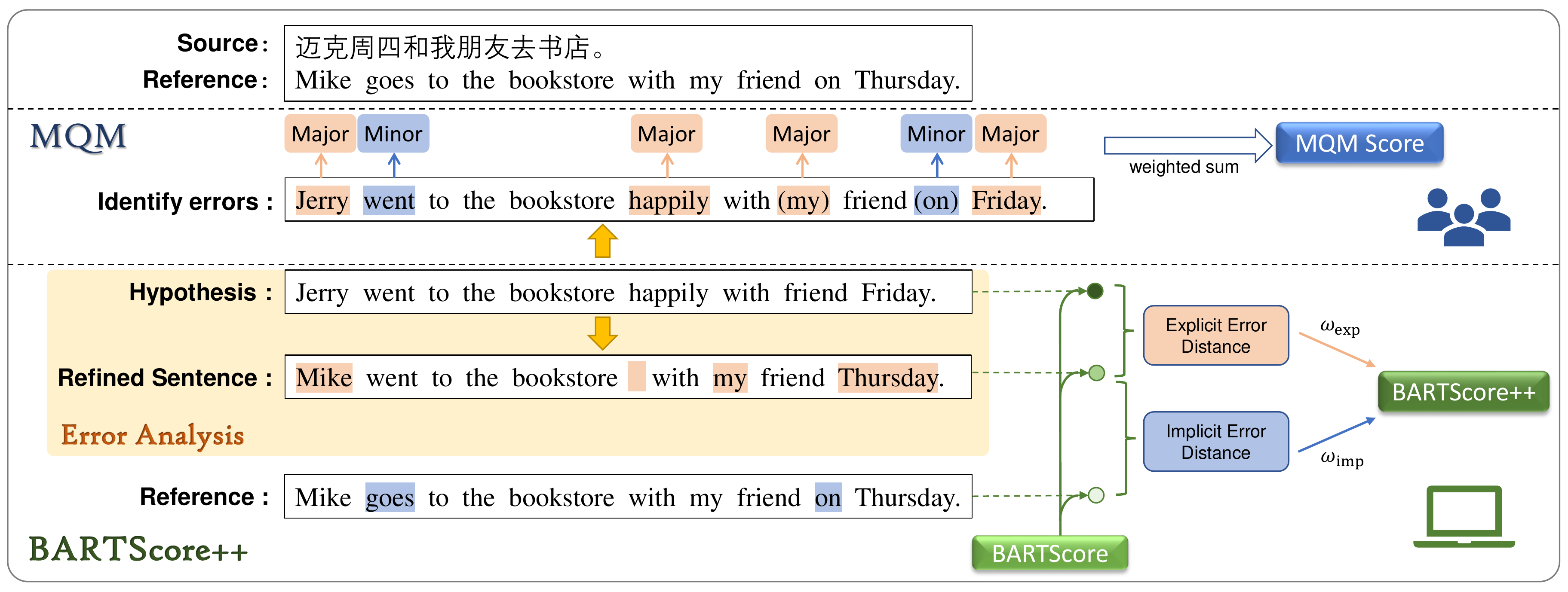}
    \caption{\textbf{An analogy between MQM and BARTScore++}. We show an evaluation example from machine translation (zh-en). \textbf{Top}: Source and reference sentence provided for evaluation. \textbf{Medium}: An annotation example using MQM framework. Errors in the hypothesis are assigned with \textit{Major} and \textit{Minor}. MQM score is computed through the weighted sum of these errors. \textbf{Bottom}: BARTScore++. The hypothesis is first refined through an error analysis framework. The refined sentence is then used to obtain the distance of explicit/ implicit errors through vanilla BARTScore. Different weights are finally assigned to these errors to get a more accurate score.}
    \label{fig:bartscore++}
\end{figure*}
\subsection{BARTScore}

BARTScore is a state-of-the-art metric proposed by \citet{yuan2021bartscore} for universal NLG evaluation. The idea of BARTScore is to utilize the generation probabilities of a large pre-trained model BART \citep{lewis-etal-2020-bart} to measure the quality of sentences. It autoregressively computes the log probabilities of each token in the hypothesis, and then averages them as the overall score. This evaluation process can be formally written as:
\begin{equation}
    \textrm{BARTScore} = \frac{1}{N}\sum_{t=1}^{N}{\log{p_{\theta}\left(y_t|y_{<t}, \boldsymbol{r}\right)}} \notag
\end{equation}
 Based on this formulation, BARTScore creates specific variants for different evaluation scenarios. We summarize their usage in Appendix~\ref{sec:appendix_bartscoreusage}. For simplification, we use the notation of $\textrm{BARTS}(\boldsymbol{y}, \boldsymbol{r})$ when vanilla BARTScore is further applied in this paper.

\begin{table*}[th]
 \centering
 \renewcommand\arraystretch{1.2}
 \small
 \setlength{\tabcolsep}{3pt}{
 \resizebox{\linewidth}{!}{\begin{tabular}{ccccccccccccc}\toprule
   \textbf{Source}: & \multicolumn{10}{l}{\begin{CJK}{UTF8}{gbsn}迈克周四和我朋友去书店。 \end{CJK}} \\
   \textbf{Reference}: & \multicolumn{10}{l}{Mike goes to the bookstore with my friend on Thursday.} \\\midrule
   \textbf{Iteration} & \multicolumn{11}{c}{\textbf{Refined Sentence}} & \textbf{BARTScore}~($\uparrow$) \\\midrule
   0 & 
   \makecell{\hl{Jerry} \\[-2pt] \tiny{\textcolor{red}{\textbf{-15.90}}}} &
   \makecell{went \\[-2pt] \tiny{-2.82}} &
   \makecell{to \\[-2pt] \tiny{-0.47}} &
   \makecell{the \\[-2pt] \tiny{-0.71}} &
   \makecell{bookstore \\[-2pt] \tiny{-2.27}} &
   \makecell{happily \\[-2pt] \tiny{-13.33}} &
   \makecell{with \\[-2pt] \tiny{-0.55}} &
   &
   \makecell{friend \\[-2pt] \tiny{-4.69}} &
   \makecell{Friday \\[-2pt] \tiny{-4.78}} &
   \makecell{. \\[-2pt] \tiny{-0.24}} &
   -3.81 \\
   1 &
   \makecell{Mike \\[-2pt] \tiny{-4.44}} &
   \makecell{went \\[-2pt] \tiny{-2.67}} &
   \makecell{to \\[-2pt] \tiny{-0.49}} &
   \makecell{the \\[-2pt] \tiny{-0.71}} &
   \makecell{bookstore \\[-2pt] \tiny{-2.30}} &
   \makecell{\hl{happily} \\[-2pt] \tiny{\textcolor{red}{\textbf{-13.21}}}} &
   \makecell{with \\[-2pt] \tiny{-0.62}} &
   &
   \makecell{friend \\[-2pt] \tiny{-4.70}} &
   \makecell{Friday \\[-2pt] \tiny{-4.82}} &
   \makecell{. \\[-2pt] \tiny{-0.23}} &
   -2.85 \\
   2 &
   \makecell{Mike \\[-2pt] \tiny{-4.44}} &
   \makecell{went \\[-2pt] \tiny{-2.67}} &
   \makecell{to \\[-2pt] \tiny{-0.49}} &
   \makecell{the \\[-2pt] \tiny{-0.71}} &
   \makecell{bookstore \\[-2pt] \tiny{-2.30}} &
   &
   \makecell{with \\[-2pt] \tiny{-0.61}} &
   &
   \makecell{\hl{friend} \\[-2pt] \tiny{\textcolor{red}{\textbf{-4.78}}}} &
   \makecell{Friday \\[-2pt] \tiny{{\textbf{-4.51}}}} &
   \makecell{. \\[-2pt] \tiny{-0.26}} &
   -1.89 \\
   3 &
   \makecell{Mike \\[-2pt] \tiny{-4.44}} &
   \makecell{went \\[-2pt] \tiny{-2.67}} &
   \makecell{to \\[-2pt] \tiny{-0.49}} &
   \makecell{the \\[-2pt] \tiny{-0.71}} &
   \makecell{bookstore \\[-2pt] \tiny{-2.30}} &
   &
   \makecell{with \\[-2pt] \tiny{-0.61}} &
   \makecell{my \\[-2pt] \tiny{-0.96}} &
   \makecell{friend \\[-2pt] \tiny{-0.10}} &
   \makecell{Thursday \\[-2pt] \tiny{-2.42}} &
   \makecell{. \\[-2pt] \tiny{-0.13}} &
   -1.24 \\\bottomrule
 \end{tabular}}}
 \caption{\textbf{An example of error analysis framework,} specifically, \textbf{detect-correct algorithm}. Scores under each token represent the log generation probability of this token obtained by vanilla BARTScore. Worse tokens detected by error analysis in each iteration are highlighted in \hl{yellow}, and their corresponding scores are in \textbf{\textcolor{red}{red}}. }
 \label{tab:refine_example}
\end{table*}

\subsection{MQM}

MQM is an error-based human evaluation framework, which is commonly agreed to be more reliable than traditional human evaluation techniques \citep{freitag-etal-2021-results}. In MQM framework, each evaluator is asked to identify all errors in a sentence and categorize them into \textit{Major} and \textit{Minor} levels indicating their severities. Sentences will be marked an \textit{Non-translation Error} if they are not possible to reliably identify errors. Major/ Minor errors are then assigned with different weights, and the final MQM score is computed through the weighted sum of errors \citep{freitag-etal-2021-experts}. Inspired by the mechanism of MQM, we take a step toward human-like evaluation by incorporating error analysis into BARTScore.

\section{BARTScore++}

To better understand how BARTScore++ works, we show an example of our method in Figure~\ref{fig:bartscore++}.

\subsection{Explicit/ Implicit Error Distance} \label{sec:errordistance}

Analogous to major errors in MQM, we define \textit{Explicit Errors} to refer to errors that can be easily identified. In our example, mistranslations of name ("Mike" $\rightarrow$ "Jerry") and date ("Thursday" $\rightarrow$ "Friday"), omission of "my", and addition of "happily" are all considered explicit errors. Analogous to minor errors, we define \textit{Implicit Errors} to indicate the semantic imperfections (e.g. disfluency, awkwardness) that may not influence the overall meanings. In our example, the replacement of "goes" with "went", and the omission of the preposition "on" are considered implicit errors because they are smaller imperfections in grammar. 

To measure the influence of Explicit/ Implicit errors in the hypothesis $\boldsymbol{y}$, we define \textit{Refined Sentence} $\boldsymbol{y}^*$ as a better hypothesis, where explicit errors are corrected. In this way, the distance of explicit/ implicit error can be computed by\footnote{We set $\textrm{BARTS}(\boldsymbol{r}) = 0$ when multiple references are provided because this makes $\boldsymbol{r}$ uncertain and causes damage to the final score in our experiments.}:
\begin{align}
    \textrm{Dist}_{\textrm{exp}} &= \textrm{BARTS}(\boldsymbol{y}^*, \boldsymbol{r}) - \textrm{BARTS}(\boldsymbol{y}, \boldsymbol{r}) \notag \\
    \textrm{Dist}_{\textrm{imp}} &= \textrm{BARTS}(\boldsymbol{r}, \boldsymbol{r}) - \textrm{BARTS}(\boldsymbol{y}^*, \boldsymbol{r}) \notag
\end{align}
We then focus on how to 1) obtain the refined sentence $\boldsymbol{y}^*$ and 2) take both explicit/ implicit errors into consideration and obtain the final score.

\subsection{Error Analysis Framework} \label{sec:revising-framework}

We introduce an automatic error analysis framework to generate the refined sentence $\boldsymbol{y}^*$ by correcting explicit errors in the hypothesis $\boldsymbol{y}$. We first adopt a simple \textbf{non-translation test} to decide whether $\boldsymbol{y}$ will be refined or not. Then, a \textbf{detect-correct algorithm} is performed iteratively, in each round one token is detected and then corrected. An example of this is shown in Table~\ref{tab:refine_example}. This algorithm repeats for a determined number of iterations $T$, where at the end of each round the refined sentence $\boldsymbol{y}^*$ is updated and becomes a new refining target. In our example, the hypothesis $\boldsymbol{y}$ is refined for $3$ times, where the mistranslated token "Jerry" is detected in Round 1, and the addition of "happily" is detected and deleted in Round 2. Afterwards, "friend" and "Friday" are detected one at a time, with the former corrected by adding "my" and the latter corrected by replacing with "Thursday".

\paragraph{Test Non-Translation Error} 
Non-Translation Error is used in MQM \citep{freitag-etal-2021-experts} to refer to the translation which is too badly garbled or is unrelated to the source. If the hypotheses contain severe problems such as off-target issues \citep{zhang-etal-2020-improving}, directly refining them will consume excessive computational cost. To avoid this problem, we run a test beforehand as a rough measure to filter these hypotheses with low quality. We consider two kinds of strategies:

\begin{itemize}
    \item[1.] \textbf{Token-level overlap ratio} w.r.t the reference. Inspired by overlap-based metrics like BLEU \citep{papineni-etal-2002-bleu} or TER \citep{snover-etal-2006-study}, the hypothesis with non-translation errors is quite different from its reference, resulting in a low overlap ratio. Since good translations like paraphrased sentences \citep{freitag-etal-2020-bleu} may not have significant overlap with the reference, we adopt the other strategy as a double-check.
    
    \item[2.] \textbf{Percentage of tokens with low generation probability}. Token-level log generation probability can be directly obtained from vanilla BARTScore as $\log{p_{\theta}\left(y_t|y_{<t}, \boldsymbol{r}\right)}$. If most tokens' generation probabilities are lower than the average score (vanilla BARTScore), we mark this sentence as non-translation. This strategy is more stable but less efficient.
    
\end{itemize}

\paragraph{Detect} In this step, we choose one token $\hat{y_t}$ with the lowest generation probabilities as the token to be corrected. This procedure can be denoted as:
\begin{equation}
    \hat{y_t} = \mathop{\arg\min}\limits_{y_t}
    \big\{ p_{\theta}\left( y_t | y_{<t}, \boldsymbol{r} \right) \big\} \notag
\end{equation}

\paragraph{Correct} In this step, we leverage the distribution of generation $p_{\theta}\left( \cdot | y_{<t}, \boldsymbol{r} \right)$ to propose several refining options from vocabulary $V$. We apply a \textbf{top-$k$ sampling} method \citep{fan-etal-2018-hierarchical} to obtain a set of candidate tokens ($\mathcal{W}$) with the highest generation probability:
\begin{equation}
    \mathcal{W} = \mathop{\arg\max}\limits_{w \in \boldsymbol{V}}
    \Big\{ p_{\theta}\left( w | y_{<t}, \boldsymbol{r} \right) , k \Big\} \notag
\end{equation}

Then, a set of refined sentences $\mathcal{S}$ is proposed. Following \citet{snover-etal-2006-study}, we apply three types of editing strategies, including insertion of a candidate token $w \in \mathcal{W}$, deletion of token $\hat{y_t}$, and substitution of $\hat{y_t}$ for a candidate token $w \in \mathcal{W}$. Finally, we use vanilla BARTScore to select the best sentence $\hat{\boldsymbol{y}}^{*}$ as the refining strategy. 
\begin{equation}
    \hat{\boldsymbol{y}}^{*} = \mathop{\arg\max}\limits_{\hat{\boldsymbol{y}} \in \mathcal{S}} \textrm{BARTS} \left(\hat{\boldsymbol{y}}, \boldsymbol{r} \right) \notag
\end{equation}
After each cycle, the hypothesis $\boldsymbol{y}$ will be temporarily replaced by $\hat{\boldsymbol{y}}^*$ and used as the input of the next iteration.

This detect-correct algorithm repeatedly detects the worst token $\hat{y_t}$ and corrects it. It starts with the original hypothesis $\boldsymbol{y}$ and ends after a constant number of edits. We also set an early-stop mechanism if the BARTScore of the refined sentence is lower than the hypothesis of the last iteration. 

In this way, we obtain the refined sentence $\boldsymbol{y}^*$, which is also a by-product of our method.

\subsection{Assigning Error Weights} \label{sec:assign-weights}

With the help of the error analysis framework, explicit errors in the hypothesis are refined, resulting in a refined sentence $\boldsymbol{y}^{*}$. We simply use a weighted sum method to achieve the final score:
\begin{equation}
    \textrm{BARTScore++} = -(\textrm{Dist}_{\textrm{exp}} \omega_{\textrm{exp}} + \textrm{Dist}_{\textrm{imp}} \omega_{\textrm{imp}}) \notag
\end{equation}
where $\omega_{\textrm{exp}}$ and $\omega_{\textrm{imp}}$ weigh the importance of explicit and implicit errors respectively.\footnote{Following the same pattern as in \citet{yuan2021bartscore}, we reverse the score to ensure BARTScore++ ranging from $-\infty$ to 0, with a higher score being a better quality of the sentence.}

\section{Experiment Setup}

\subsection{Tasks and Datasets}

\begin{table*}[h]
 \centering
 \renewcommand\arraystretch{1.2}
 \small
 \setlength{\tabcolsep}{4pt}{
 \resizebox{\linewidth}{!}{\begin{tabular}{lcccccccccccc}\toprule
       \multirow{2}{*}{\textbf{Metrics}} & \multicolumn{6}{c}{\textbf{High-Resource}} & \multicolumn{6}{c}{\textbf{Low-Resource}}
         \\ \cmidrule(lr){2-7} \cmidrule(lr){8-13}
         & \textbf{cs} & \textbf{de} & \textbf{ja} & \textbf{ru} & \textbf{zh} & \textbf{Avg.} & \textbf{iu} & \textbf{km} & \textbf{pl} & \textbf{ps} & \textbf{ta} & \textbf{Avg.} \\\midrule
    \textit{Supervised Baselines} \\\hdashline[3pt/3pt]
    BLEURT & \underline{12.97} & 6.61 & 12.82 & 6.55 & 11.62 & \cellcolor{myblue}{10.12} & 26.78 & 31.09 & 2.76 & \underline{18.05} & 16.88 & \cellcolor{myblue}19.11 \\
    COMET & 11.02 & \underline{9.04} & 12.47 & \underline{12.07} & \underline{14.50} & \cellcolor{myblue}{11.82} & 27.19 & 29.84 & 9.90 & 15.71 & 15.81 & \cellcolor{myblue}19.69 \\\midrule
    \textit{Unsupervised Baselines} \\\hdashline[3pt/3pt]
    BLEU & 3.90 & -2.93 & 7.00 & -3.47 & 6.39 & \cellcolor{myblue}2.18 & 15.41 & 22.72 & -5.25 & 10.47 & 7.19 & \cellcolor{myblue}10.11 \\
    BERTScore & 11.60 & 4.03 & 12.85 & 5.21 & 10.58 & \cellcolor{myblue}8.85 & 24.74 & 30.01 & 2.78 & 14.29 & 13.41 & \cellcolor{myblue}17.04 \\
    PRISM & 12.42 & 2.67 & 13.46 & 7.22 & 11.65 & \cellcolor{myblue}9.48 & 25.37 & 30.44 & 5.70 & \textbf{16.51} & 14.78 & \cellcolor{myblue}18.56 \\\midrule
    \textit{BARTScore} \\\hdashline[3pt/3pt]
    Vanilla BARTScore & 11.81 & 5.55 & 13.62 & 9.22 & 13.12 & \cellcolor{myblue}10.66 & 26.93 & 32.27 & 7.64 & 15.54 & 16.63 & \cellcolor{myblue}19.80 \\
    + Prompt & 12.31 & 7.26 & 14.16 & 11.13 & 13.13 & \cellcolor{myblue}11.60 & 27.11 & 32.16 & 9.44 & 16.05 & 16.05 & \cellcolor{myblue}20.32 \\\midrule
    \textit{Ours - BARTScore++} \\\hdashline[3pt/3pt]
    + \textbf{Error Analysis} & 12.06 & 7.23\ddag & 15.08\ddag & 9.98\ddag & 13.32\ddag & \cellcolor{myblue}11.54 & 27.37\dag & \underline{\textbf{32.38}}\dag & 8.44\ddag & 15.94 & 17.09\ddag & \cellcolor{myblue}20.24 \\
    + Prompt + \textbf{Error Analysis} & \textbf{12.65}\dag & \textbf{8.75}\ddag & \underline{\textbf{15.40}}\ddag & \textbf{11.76}\ddag & \textbf{13.35}\ddag & \cellcolor{myblue}\underline{\textbf{12.38}} & \underline{\textbf{27.60}}\ddag & 32.33\dag & \underline{\textbf{10.14}}\ddag & 16.40 & \underline{\textbf{17.39}}\ddag & \cellcolor{myblue}\underline{\textbf{20.77}}
    \\\bottomrule
 \end{tabular}}}
 \caption{\textbf{Segment-level Kendall's $\tau$ correlation} (\%) results on English-targeted language pairs of \textbf{WMT20 Metrics Shared Task} test set. \textbf{Bold} and \underline{Underlined} values refer to the best result among unsupervised metrics and all metrics, respectively. $\dag$ indicates BARTScore++ significantly outperforms BARTScore without error analysis, and $\ddag$ indicates BARTScore++ further significantly outperform other unsupervised baselines. }
 \label{tab:mt-results}
\end{table*}

\paragraph{Tasks} We follow \citet{yuan2021bartscore} to consider three different tasks: summarization (SUM), machine translation (MT), and data-to-text (D2T).
\paragraph{Datasets for Translation} We obtain 
the machine-translated texts and reference texts from the \texttt{WMT20} metrics shared task \citep{mathur2020results}. We use the DARR corpus and consider 10 language pairs, which are \texttt{cs-en}, \texttt{de-en}, \texttt{ja-en}, \texttt{ru-en}, \texttt{zh-en}, \texttt{iu-en}, \texttt{km-en}, \texttt{pl-en}, \texttt{ps-en}, \texttt{ta-en}. For human evaluation,  we also consider Multidimensional Quality Metric (MQM) for \texttt{zh-en} provided by \citet{freitag-etal-2021-experts} in \S\ref{sec:analysis}, comprising human evaluations of 8 best-performing systems of \texttt{WMT20} (excluding human translations).

\paragraph{Datasets for Summarization} (1) \texttt{REALSumm} \citep{bhandari-etal-2020-evaluating} is a meta-evaluation dataset for text summarization which measures pyramid-recall of each system-generated summary. (2) \texttt{SummEval} \citep{fabbri-etal-2021-summeval} is a collection of human judgments of model-generated summaries on the \texttt{CNNDM} dataset annotated by both expert judges and crowd-source workers. Each system-generated summary is gauged through the lens of coherence, factuality, fluency, and informativeness. (3) \texttt{NeR18} The \texttt{NEWSROOM} dataset \citep{grusky-etal-2018-newsroom} contains 60 articles with summaries generated by 7 different methods are annotated with human scores in terms of coherence, fluency, informativeness, relevance. 
\paragraph{Datasets for Factuality} (1) \texttt{Rank19} \citep{falke-etal-2019-ranking} is used to meta-evaluate factuality metrics. It is a collection of 373 triples of a source sentence with two summary sentences, one correct and one incorrect. (2) \texttt{QAGS20} \citep{wang2020qags} collected 235 test outputs on \texttt{CNNDM} dataset from \citet{gehrmann-etal-2018-bottom} and 239 test outputs on \texttt{XSUM} dataset \citep{narayan-etal-2018-dont} from BART fine-tuned on \texttt{XSUM}. Each summary sentence is annotated with correctness scores w.r.t. factuality.
\paragraph{Datasets for Data-to-Text} We consider the following datasets which target utterance generation for spoken dialogue systems. (1) \texttt{BAGEL} \citep{mairesse-etal-2010-phrase} provides information about restaurants. (2) \texttt{SFHOT} \citep{wen-etal-2015-semantically} provides information about hotels in San Francisco. (3) \texttt{SFRES} \citep{wen-etal-2015-semantically} provides information about restaurants in San Francisco. They contain 202, 398, and 581 samples respectively, each sample consists of one meaning representation, multiple references, and utterances generated by different systems.

\subsection{Baselines and Meta-evaluation}

\paragraph{Baselines} We compare our method with several commonly used baseline metrics for evaluating text generation, including BLEU \citep{papineni-etal-2002-bleu}, BERTScore \citep{zhang2020bertscore}, MoverScore \citep{zhao-etal-2019-moverscore} and PRISM \citep{thompson-post-2020-automatic}. For MT task, we also consider supervised metrics that leverage human judgments to train, including COMET \citep{rei-etal-2020-comet} and BLEURT \citep{sellam-etal-2020-bleurt}. For factuality evaluation on the summarization task, we compare BARTScore++ with the best-performing factuality metrics FactCC \citep{kryscinski2020evaluating} and QAGS \citep{wang2020qags}. We reproduce BARTScore and their variants using the official codes released by \citet{yuan2021bartscore}.

\paragraph{Meta-evaluation} For the meta-evaluation of all experiments, we follow the same settings as \citet{yuan2021bartscore}. Specifically, We apply \textit{Kendall's $\tau$} for MT task to measure the correlation of metrics with human evaluation.\footnote{Since the meta-evaluation method is very sensitive to outliers (systems whose scores are far away from the rest of the systems) \citep{mathur-etal-2020-tangled}, we remove these outlier systems when computing correlations.} For SUM and D2T tasks, we use \textit{Spearman correlation} except for the Rank19 dataset, where \textit{Accuracy} is used to measure the percentage of correct ranking between factual texts and non-factual texts.

\paragraph{Significance Tests} To perform rigorous analysis, we follow \citet{yuan2021bartscore} to adopt the bootstrap method (p-value < 0.05) \citep{koehn-2004-statistical} for pair-wise significance tests.

\subsection{Setup} \label{sec:setup}

\paragraph{Settings of BARTScore} As for the backbone BART, we use the same settings in BARTScore \citep{yuan2021bartscore} for specific tasks, including BART-large, BART-CNN (fine-tuned on \texttt{CNNDM}) and BART-CNN-PARA (further fine-tuned on \texttt{ParaBank2}). We also perform the same prompting strategy as in BARTScore \citep{yuan2021bartscore}. We report the detail settings in Appendix \ref{sec:appendix_bartscoreusage}.

\paragraph{Hyper Parameters} We set $k=10$ when applying the top-$k$ sampling method to find candidate tokens. For a fair comparison, we set the batch size to 4 for all experiments.

\section{Experimental Results} \label{sec:expresult}

\begin{table*}[ht]
 \centering
 \renewcommand\arraystretch{1.2}
 \small
 \setlength{\tabcolsep}{4pt}{
 \resizebox{\linewidth}{!}{\begin{tabular}{lcccccccccr}\toprule
    \multirow{2}{*}{\textbf{Metrics}} & \textbf{REALSumm} &  \multicolumn{4}{c}{\textbf{SummEval}} & \multicolumn{4}{c}{\textbf{NeR18}} & 
         \\ \cmidrule(lr){2-2} \cmidrule(lr){3-6} \cmidrule(lr){7-10}
         & \textbf{COV} & \textbf{COH} & \textbf{FAC} & \textbf{FLU} & \textbf{INFO} & \textbf{COH} & \textbf{FLU} & \textbf{INFO} & \textbf{REL} & \textbf{Avg.} \\\midrule
    \textit{Baselines} \\\hdashline[3pt/3pt]
    ROUGE & \textbf{49.75} & 16.68 & 15.96 & 11.50 & 32.64 & 9.46 & 10.36 & 13.04 & 14.73 & \cellcolor{myblue}19.35 \\
    BERTScore & 44.04 & 28.38 & 10.97 & 19.26 & 31.20 & 14.75 & 17.03 & 13.09 & 16.34 & \cellcolor{myblue}21.67 \\
    MoverScore & 37.24 & 15.91 & 15.71 & 12.86 & 31.77 & 16.15 & 11.97 & 18.80 & 19.54 & \cellcolor{myblue}19.99 \\
    PRISM & 41.10 & 24.88 & 34.52 & 25.36 & 21.16 & 57.28 & 53.20 & 56.13 & 55.34 & \cellcolor{myblue}41.00 \\\midrule
    \textit{BARTScore} \\\hdashline[3pt/3pt]
    Vanilla BARTScore & 47.42 & \textbf{44.67} & 38.11 & 35.64 & 35.53 & 67.89 & 67.00 & 64.67 & 60.51 & \cellcolor{myblue}51.27 \\
    + Prompt & 48.71 & 40.75 & 37.76 & 33.74 & 36.89 & 70.14 & 67.89 & 68.60 & 62.04 & \cellcolor{myblue}51.83 \\\midrule
    \textit{Ours - BARTScore++} \\\hdashline[3pt/3pt]
    + \textbf{Error Analysis} & 47.76 & \textbf{44.67}\dag & \textbf{38.48}\dag & \textbf{35.66}\dag & 35.53\dag & 68.62\ddag & 67.79\dag & 68.60\ddag & 61.15\ddag & \cellcolor{myblue}51.73 \\
    + Prompt + \textbf{Error Analysis} & 49.00 & 40.83\dag & 38.08\dag & 33.88\dag & \textbf{37.01}\dag & \textbf{70.44}\ddag & \textbf{68.75}\ddag & \textbf{69.66}\ddag & \textbf{63.04}\ddag & \cellcolor{myblue}\textbf{52.30}
    \\\bottomrule
 \end{tabular}}}
 \caption{\textbf{Spearman correlation} (\%) results on three \textbf{text summarization datasets}. Best viewed in \textbf{Bold}. $\dag$ and $\ddag$ indicate BARTScore++ significantly outperforms all baselines and powerful BARTScore without error analysis, respectively. }
 \label{tab:sum-results}
\end{table*}

\begin{table}[h]
 \centering
 \small
 \renewcommand\arraystretch{1.2}
 \setlength{\tabcolsep}{3pt}{
 \resizebox{\linewidth}{!}{\begin{tabular}{lcccl}\toprule
         \multirow{2}{*}{\textbf{Metrics}} & \textbf{Rank19} & \textbf{Q-CNN} & \textbf{Q-XSUM}
         \\\cmidrule(lr){2-2} \cmidrule(lr){3-4}
         & \textbf{Acc.(\%)} & \multicolumn{2}{c}{\textbf{Pearson(\%)}} \\\midrule
    \textit{Baselines} \\\hdashline[3pt/3pt]
    ROUGE & 63.00 & 45.91 & 9.70 \\
    BERTScore & 71.31 & 57.60 & 2.38 \\
    MoverScore & 71.31 & 41.41 & 5.41 \\
    PRISM & 78.02 & 47.87 & 2.50 \\\midrule
    \textit{Factuality Metrics} \\\hdashline[3pt/3pt]
    FactCC & 70.00 & - & - \\
    QAGS & 71.20 & 54.50 & 17.50 \\
    Human & 83.90 & - & - \\\midrule
    \textit{BARTScore} \\\hdashline[3pt/3pt]
    Vanilla BARTScore & 83.65 & 73.47 & 18.38 \\
    + Prompt & 79.62 & 71.85 & 9.40
    \\\midrule
    \textit{Ours - BARTScore++} \\\hdashline[3pt/3pt]
    + \textbf{Error Analysis} & \textbf{84.18}\dag & \textbf{73.97}\ddag & \textbf{19.33}\ddag \\
    + Prompt + \textbf{Error Analysis} & 80.70\ddag & 72.60\ddag & 10.55 
    \\\bottomrule
 \end{tabular}}}
 \caption{\textbf{Results on Factuality Datasets}, where "Q" is short for QAGS. $\dag$ and $\ddag$ indicate BARTScore++ significantly outperforms all baselines and powerful BARTScore without error analysis, respectively. }
 \label{tab:fac-results}
\end{table}

\subsection{Machine Translation}

Table~\ref{tab:mt-results} shows segment-level Kendall $\tau$ correlation of metrics on \texttt{WMT20}. We can observe that BARTScore++ can achieve state-of-the-art performance on all language pairs (most significantly outperform vanilla BARTScore except \texttt{ps-en}). The average correlation of BARTScore can surpass all supervised and unsupervised metrics by a large margin in both high-resource and low-resource scenarios. This confirms our intuition that with analysis of explicit/implicit errors, BARTScore++ will agree more with human evaluations compared with vanilla BARTScore.

Regarding the prompting strategy, we also observe that 1) our proposed error analysis mechanism in BARTScore++ can achieve a similar amount of correlation improvement as that of prompting, and 2) incorporating both prompting and error analysis can further push SOTA results, confirming the orthogonality of error analysis and prompting strategies upon BARTScore.

\subsection{Text Summarization}

Results on \texttt{REALSumm}, \texttt{SummEval} and \texttt{NeR18} are showed in Table~\ref{tab:sum-results}. We observe that: 
1) BARTScore++ surpasses all other metrics including BARTScore variants for all test settings except \texttt{REALSumm}. In most aspects, our purposed method can significantly improve the performance of vanilla BARTScore; 2) Also, error analysis mechanism in BARTScore++ can achieve a similar amount of correlation improvement as that of prompting, which testify the importance of considering errors in summarization evaluation. 

\begin{table}[t]
 \centering
 \small
 \renewcommand\arraystretch{1.2}
 \setlength{\tabcolsep}{1pt}{
 \resizebox{\linewidth}{!}{\begin{tabular}{lcccl}\toprule
    \textbf{Metrics} & \textbf{BAGEL} & \textbf{SFRES} & \textbf{SFHOT} & \textbf{Avg.} \\\midrule
    \textit{Baselines} \\\hdashline[3pt/3pt]
    ROUGE & 23.43 & 11.57 & 11.75 & \cellcolor{myblue}15.58 \\
    BERTScore & 28.91 & 15.64 & 13.54 & \cellcolor{myblue}19.36 \\
    MoverScore & 28.37 & 15.27 & 17.23 & \cellcolor{myblue}20.29 \\
    PRISM & 30.49 & 15.47 & 19.64 & \cellcolor{myblue}21.87 \\\midrule
    \textit{BARTScore} \\\hdashline[3pt/3pt]
    Vanilla BARTScore & 31.89 & 19.52 & 21.65 & \cellcolor{myblue}24.35 \\
    + Prompt & 33.28 & 23.74 & 23.81 & \cellcolor{myblue}26.94 \\\midrule
    \textit{Ours - BARTScore++} \\\hdashline[3pt/3pt]
    + \textbf{Error Analysis} & 32.67\dag & 19.74\dag & 25.63\ddag & \cellcolor{myblue}26.00 \\
    + Prompt + \textbf{Error Analysis} & \textbf{34.12}\ddag & \textbf{23.99}\ddag & \textbf{26.04}\ddag & \cellcolor{myblue}\textbf{28.02}
    \\\bottomrule
 \end{tabular}}}
 \caption{\textbf{Spearman correlation} (\%) of different metrics over three \textbf{Data-to-Text datasets}. ``$\dag$'' and ``$\ddag$'' indicate that BARTScore++ significantly outperforms all baselines and powerful BARTScore without error analysis, respectively.}
 \label{tab:d2t-results}
\end{table}

\paragraph{Analysis on factuality datasets} As shown in Table~\ref{tab:fac-results}, we observe that BARTScore++ significantly outperforms all other metrics on all three datasets. Strikingly, BARTScore++ can even surpass human baseline on \texttt{Rank19} by a large margin. This further confirms the universality of our proposed error analysis strategy.

\subsection{Data-to-Text} 

Results on Data-to-Text datasets are shown in Table~\ref{tab:d2t-results}. We can see that BARTScore++ can again surpass existing methods and significantly outperform vanilla BARTScore on \texttt{SFRES} and \texttt{SFHOT}. We further find weights on explicit errors are consistently larger than implicit errors on all three datasets. This suggests that we should focus more on explicit errors when evaluating Data-to-Text tasks.

\section{Analysis} \label{sec:analysis}

To better understand the mechanism by which BARTScore++ achieves these promising results, we take a closer look at BARTScore++ by testing their correlation with MQM human evaluations, an error-based evaluation framework annotated by human experts \citep{freitag-etal-2021-experts} and answer three questions: 
\begin{itemize}
    \item[\bf Q1:] How reliable is our BARTScore++ when evaluating top-performing systems?
    \item[\bf Q2:] How do explicit/ implicit error weights influence the accuracy of BARTScore++?
    \item[\bf Q3:] How does error analysis make BARTScore++ more human-like?
\end{itemize}

\subsection{BARTScore++ is Reliable When Evaluating Top-$K$ Systems}\label{sec:topk-results}

Previous studies have shown that most metrics are unreliable for evaluating best-performing systems, showing a sharp degradation of correlation with human evaluation \citep{mathur-etal-2020-tangled}. 

\begin{figure}[t]
    \centering
    \includegraphics[scale=0.45]{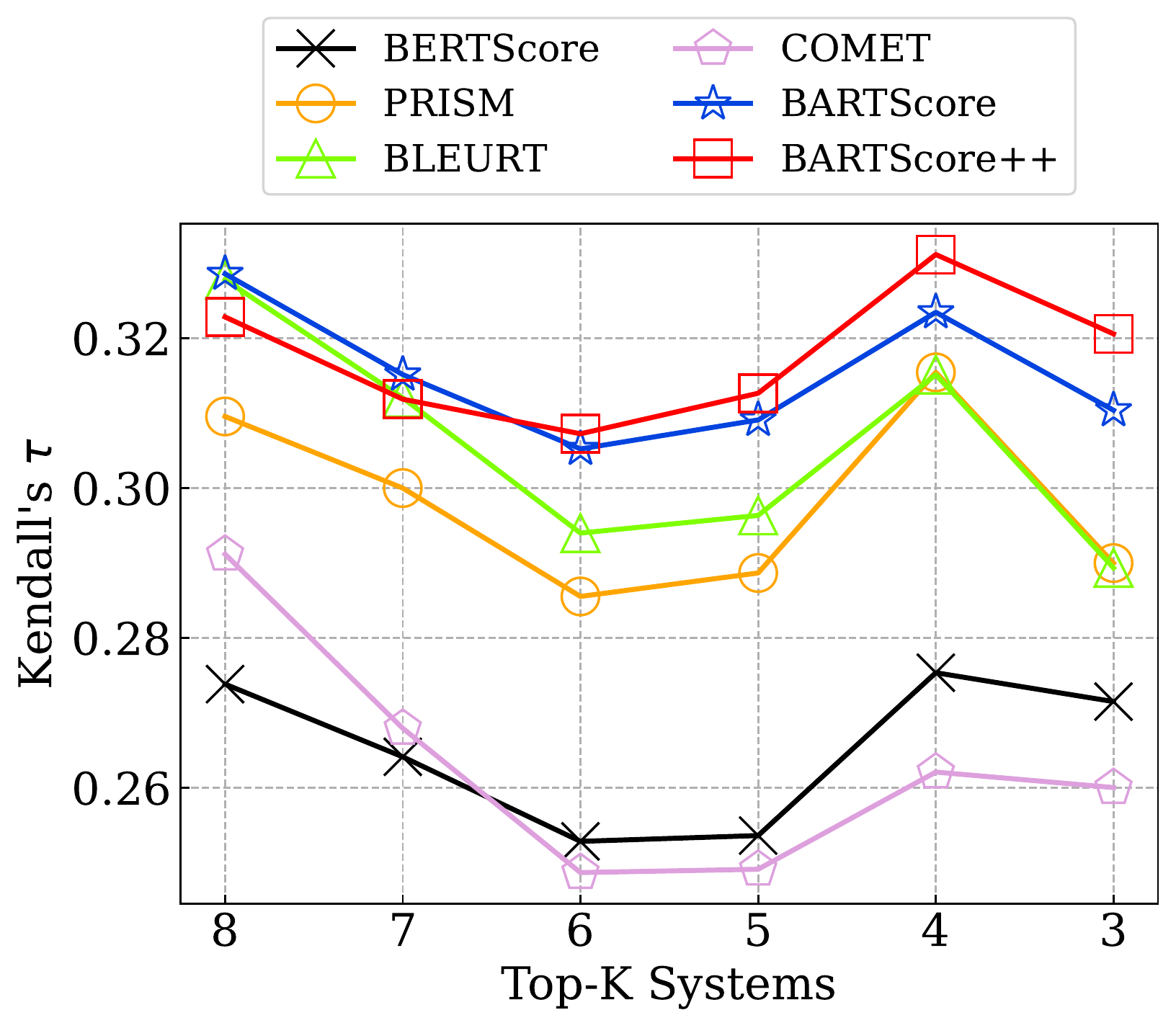}
    \caption{\textbf{Kendall correlation} of different metrics on \textbf{Top-$K$ MT systems according to MQM human evaluation dataset.}}
    \label{fig:topk}
\end{figure}

To answer {\bf Q1}, we assess our method shown in Figure~\ref{fig:topk} with several baseline metrics on Top-$K$ MT systems by computing Kendall's $\tau$ respectively .\footnote{Note that the explicit/implicit error weight is fixed to $= 1.4:1$ for fair comparison.} We can see that BARTScore++ can further improve BARTScore's performance by a large margin, especially when evaluating top-performing systems ($K < 6$). This verifies the reliability of our purposed method.

\subsection{BARTScore++ is Stable When Adjusting Error Weights}
\begin{figure}[h]
    \centering
    {{\includegraphics[scale=0.26]{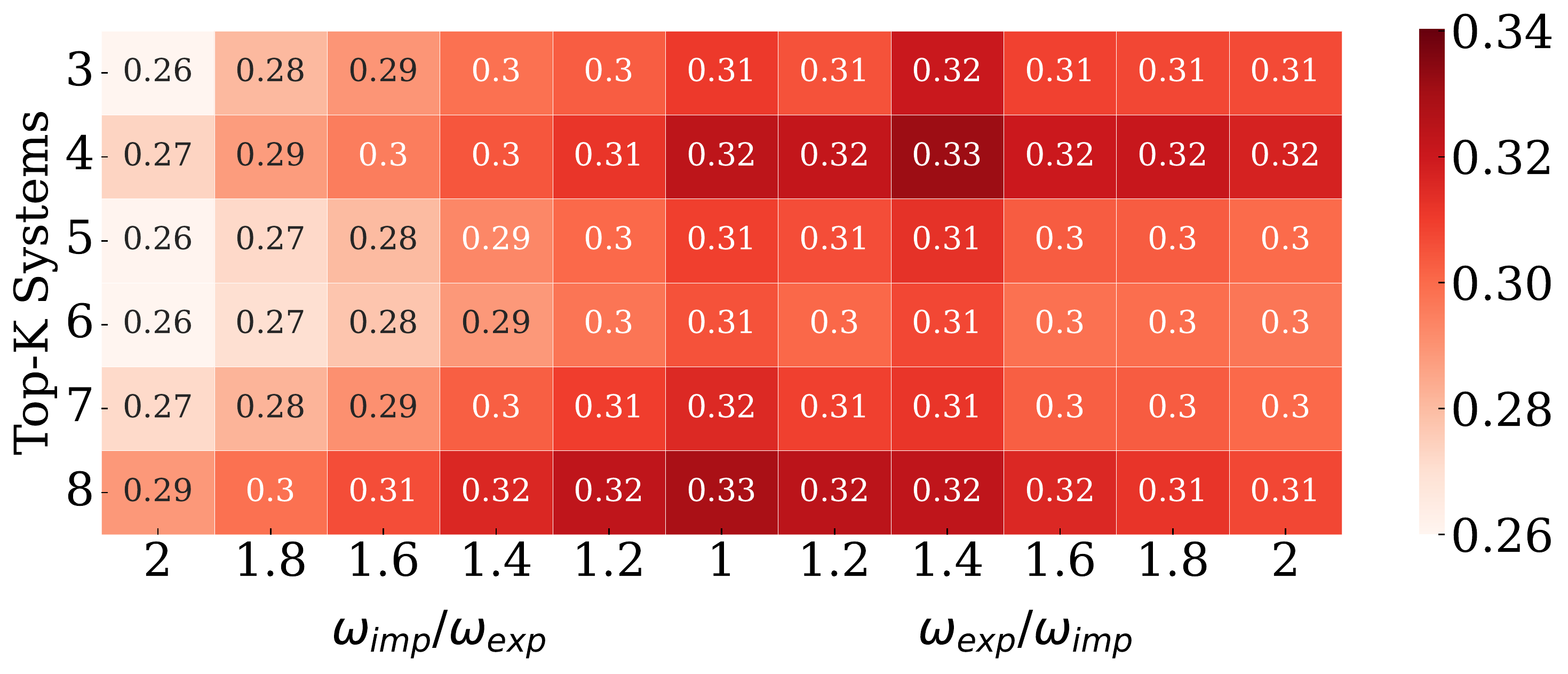} }} 
    \caption{\textbf{Effect of different Error Weights} in our BARTScore++ on top-$K$ MT systems.}
    \label{fig:finegrained}
\end{figure}

To answer {\bf Q2}, we present an analysis on adjusting hyperparameters of BARTScore++. In Figure~\ref{fig:finegrained}, as the number of systems $K$ decreases, the ratio of error weights according to the best performing BARTScore++ is fluctuating from $1$ to $1.4$. This suggests the importance of adjusting the error weights according to the overall qualities of MT systems before evaluating them. 

\subsection{BARTScore++ is More Human-Like on Discriminating Errors }

To answer {\bf Q3}, we perform a case study in Appendix~\ref{sec:appendix_casestudy} to further show the advantage of our error analysis strategies incorporated in BARTScore++. 

In Table~\ref{tab:case-study}, we can see that human evaluators consistently assign low MQM scores to explicit errors (e.g. mistranslation of "delivery" in WeChat AI in example 1, mistranslation of "yuan" in VolcTrans in example 2), but BARTScore produces contrary judgments, ignoring these errors that should be punished strictly. Through our proposed error analysis, BARTScore++ becomes more discriminative on explicit errors and reaches agreement with human judgments, while BARTScore fails to discriminate against such errors.

\section{Related Work}

\paragraph{Automatic Metrics} Automatic Evaluation Metrics are of crucial importance to the development of NLG systems. Recent research has shown great success in language model-based metrics, which can significantly outperform traditional surface-based metrics such as BLEU \citep{papineni-etal-2002-bleu}. For example, BERTScore \citep{zhang2020bertscore} and MoverScore \citep{zhao-etal-2019-moverscore} leverage contextual embeddings to measure semantic distance between reference and hypothesis. COMET \citep{rei-etal-2020-comet} and BLEURT \citep{sellam-etal-2020-bleurt} rely on human evaluations to train. In this paper, we choose BARTScore \citep{yuan2021bartscore} as the testbed because of its SOTA performance and universality on NLG tasks. Note that our error analysis strategies can also be extended to other metrics, such as PRISM \citep{thompson-post-2020-automatic}.  

\paragraph{Human Evaluation} Human evaluation techniques, such as Direct Assessment \citep{graham2017can}, are often served as `golden standard'. However, there is increasing evidence that inadequate evaluation will lead to wrong decisions \citep{toral-2020-reassessing}. This motivates elaborate evaluation proposals \citep{popovic-2020-informative, gladkoff2021hope} and MQM is one of these methodologies, grounded in explicit error analysis \citep{freitag-etal-2021-experts}. In this work, We extend error analysis strategies to BARTScore, making it trigger more human-like judgments.

\paragraph{Error Analysis} Existing automatic metrics tend to simplify the error detection procedure, such as edit distance in TER \citep{snover-etal-2006-study} and mismatch in BERTScore \citep{zhang2020bertscore}. In this work, we try to reveal the token-level judgments in BARTScore and analyze these errors through refining operations, making metrics "think" like a human, and providing more accurate evaluations. Our error analysis framework functionalizes like token-level quality estimation \citep{specia-etal-2021-findings} or automatic post-editing \citep{freitag-etal-2019-ape}. With the reference signal provided, our proposed method is more accurate and universal for NLG evaluation.

\section{Conclusion}

In this paper, we present an automatic metric BARTScore++ for evaluating natural language generation. Inspired by the advanced human evaluation MQM, BARTScore++ incorporates error analysis strategies to give a comprehensive score considering explicit and implicit errors. Experimental results show that our proposed method consistently improves the performance of BARTScore and achieves competitive results on a broad range of tasks. Our work is an early step toward human-like evaluation for automatic metrics, and we hope our BARTScore++ can motivate researchers working on NLG evaluation to focus more on human evaluation procedures such as error analysis. 

\section*{Limitations}

Limitations of BARTScore++ are three-fold:

\begin{itemize}
    \item In \S\ref{sec:errordistance}, we propose Explicit/Implicit errors to better distinguish different types of errors in generated texts. However, explicit errors only contain token-level errors that can be detected and corrected by error analysis, not involving all error types mentioned in MQM (e.g. severe fluency errors). We hope future studies can take these situations into account.
    \item In \S\ref{sec:revising-framework} we can see that our proposed error analysis framework fully relies on the generation probabilities of BART to decide how to refine the hypothesis. Still, we see that this framework may lead to false judgments due to unfaithful content. Further research can explore how to calibrate the pre-trained models during error analysis.
    \item In \S\ref{sec:assign-weights} we integrate the distance of explicit and implicit errors by simply computing their weighted sum. This can be improved by considering more factors, e.g. the overall quality of the generated text, refining iterations, and external signals. We will leave the exploration of combining these factors and designing better weighting schemes as future work.
\end{itemize}

\bibliography{anthology,arxiv_version}
\bibliographystyle{acl_natbib}

\appendix

\section{Variants of Vanilla BARTScore} \label{sec:appendix_bartscoreusage}

\paragraph{BARTScore Variants} We summarize variants of BARTScore in Table~\ref{tab:bartscore-variants}. $\mathcal{F}$ score is applied for Machine Translation and Data-to-Text tasks; recall-based BARTScore is applied in \texttt{REALSumm} due to recall-based pyramid human evaluation; BARTScore on faithfulness is applied to other summarization tasks. In our experiments, we follow the same settings as in BARTScore \citep{yuan2021bartscore}. 

\begin{table}[h]
 \centering
 \renewcommand\arraystretch{1.2}
 \small
 \setlength{\tabcolsep}{3pt}{
 \begin{tabular}{cc}\toprule
        \textbf{Variants} & \textbf{Computation using BARTScore} \\\midrule
        $\mathcal{F}$ score & $ \left(\textrm{BARTScore}_{\boldsymbol{r} \rightarrow \boldsymbol{h}} + \textrm{BARTScore}_{\boldsymbol{h} \rightarrow \boldsymbol{r}} \right) / 2 $ \\
        Recall & $\textrm{BARTScore}_{\boldsymbol{h} \rightarrow \boldsymbol{r}}$ \\
        Faithfulness & $\textrm{BARTScore}_{\boldsymbol{s} \rightarrow \boldsymbol{h}}$
    \\\bottomrule
 \end{tabular}}
 \caption{BARTScore variants and their computation method. The source, reference sentence and hypothesis are denoted as $\boldsymbol{s}$, $\boldsymbol{r}$, $\boldsymbol{h}$ respectively.}
 \label{tab:bartscore-variants}
\end{table}

\paragraph{Prompt Design} Prompting is a parameter-free method to elicit more accurate results by combining texts with a set of short phrases (prompts). BARTScore applies this method through two basic approaches: suffixing prompts on the encoder or prefixing prompts on the decoder of BART \citep{lewis-etal-2020-bart}. If multiple prompts are provided, the final BARTScore of a hypothesis is computed by averaging the score of all its generation scores using different prompts. When vanilla BARTScore is used in our method, we perform the same prompting strategy as in BARTScore \citep{yuan2021bartscore}.

\section{Case Study} \label{sec:appendix_casestudy}

\begin{table*}[ht]
 \centering
 \small
 \renewcommand\arraystretch{1.2}
 \newcommand{\tabincell}[2]{\begin{tabular}{@{}#1@{}}#2\end{tabular}}
 \setlength{\tabcolsep}{5pt}{
 \begin{tabularx}{\textwidth}{lXccc}\toprule
         \multicolumn{5}{c}{\textbf{Example 1}} \\\midrule
    Source & \multicolumn{4}{l}{\begin{CJK}{UTF8}{gbsn}9月25日，北京大兴国际机场\colorbox{green}{投运}仪式\colorbox{yellow}{隆重}举行。\end{CJK}} \\
    Reference & \multicolumn{4}{l}{On September 25th, a\colorbox{yellow}{grand}\colorbox{green}{opening}ceremony was held for the Beijing Daxing International Airport.}
    \\\midrule
    \textbf{MT system} & \textbf{Output} & \textbf{BARTScore} & \textbf{BARTScore++} & \textbf{MQM} \\\cmidrule(lr){1-5}
    VolcTrans & On September 25, the\colorbox{green}{commissioning}ceremony of Beijing Daxing International Airport was held\colorbox{yellow}{ceremoniously}. & -1.544 & \textbf{-0.827} & \textbf{-2.000} \\
    WeChat AI & On September 25, the\colorbox{yellow}{$\ \ $} \colorbox{green}{delivery} ceremony of Beijing Daxing International Airport was held. & \textbf{-1.375} & -0.855 & -5.333 \\\midrule
    \textbf{MT system} & \textbf{Revised output} & \textbf{BARTScore} & $\textrm{\textbf{Dist}}_{\textrm{\textbf{exp}}}$ & $\textrm{\textbf{Dist}}_{\textrm{\textbf{imp}}}$ \\\cmidrule(lr){1-5}
    VolcTrans & On September 25, the\colorbox{green}{commissioning}ceremony of Beijing Daxing International Airport was held\colorbox{yellow}{ceremoniously}. & -1.544 & 0.000 & 0.827 \\
    WeChat AI & On September 25, the\colorbox{yellow}{grand}\colorbox{green}{opening}ceremony for Beijing Daxing International Airport was held. & -0.884 & 0.491 & 0.168
    \\\bottomrule
 \end{tabularx}
 \begin{tabularx}{\textwidth}{lXccc}\toprule
         \multicolumn{5}{c}{\textbf{Example 2}} \\\midrule
    Source & \multicolumn{4}{l}{\begin{CJK}{UTF8}{gbsn}中国将实施更加积极主动的开放战略，创造更全面、更深入、更\colorbox{yellow}{多元}的对外开放格局，...\end{CJK}} \\
    Reference & \multicolumn{4}{l}{\tabincell{l}{China will follow a more proactive opening-up strategy, work to create a more comprehensive,\colorbox{yellow}{diverse}, and \\ deeper layout for opening-up, ...}}
    \\\midrule
    \textbf{MT system} & \textbf{Output} & \textbf{BARTScore} & \textbf{BARTScore++} & \textbf{MQM} \\\cmidrule(lr){1-5}
    VolcTrans & China will implement a more proactive opening-up strategy, create a more comprehensive, more in-depth and \colorbox{yellow}{more yuan}pattern of opening up to the outside world, ... & \textbf{-2.324} & -2.091 & -6.367 \\
    WeChat AI & China will implement a more proactive opening strategy, create a more comprehensive, deeper and \colorbox{yellow}{diversified} pattern of opening up to the outside world, ... & -2.339 & \textbf{-2.085} & \textbf{-2.067} \\\midrule
    \textbf{MT system} & \textbf{Revised output} & \textbf{BARTScore} & $\textrm{\textbf{Dist}}_{\textrm{\textbf{exp}}}$ & $\textrm{\textbf{Dist}}_{\textrm{\textbf{imp}}}$ \\\cmidrule(lr){1-5}
    VolcTrans & China will implement a more proactive opening-up strategy, work to create a more comprehensive, more \colorbox{yellow}{diverse}-depth and more deeper layout of opening up, ... & -1.429 & 0.895 & 0.838 \\
    WeChat AI & China will implement a more proactive opening up strategy, work to create a more comprehensive, diverse and more \colorbox{yellow}{diversified} layout of opening up, ... & -1.497 & 0.842 & 0.907
    \\\bottomrule
 \end{tabularx}
 }
 \caption{Two examples from WMT20 zh-en test dataset with a disagreement between BARTScore and BARTScore++ . \textbf{Bold} values refer to better segment judged by different metrics for each sample. Explicit errors and their corresponding token from other sentences are highlighted in \colorbox{yellow}{yellow} and \colorbox{green}{green}. }
 \label{tab:case-study}
\end{table*}

We also show two evaluation examples of machine translation in Table~\ref{tab:case-study} to further explain how error analysis makes BARTScore++ more human-like. We show generated texts, MQM scores, and Explicit/Implicit error distance of BARTScore++ from two out-performing systems, VolcTrans and WeChat AI, on WMT20. 

\paragraph{Example 1} We can see that WeChat AI produces a mistranslation error and an omission error (marked in \colorbox{green}{green} and \colorbox{yellow}{yellow} respectively). However, vanilla BARTScore seems to "ignore" this error and give a higher score than a better translation by VolcTrans. BARTScore++ applies an error analysis and gives a more discriminative evaluation by detecting these errors and recording them as explicit errors. Explicit error distances for both translations (0.491 vs. 0.000) are enlarged by a larger error weight, resulting in an agreement with human judgment. 

\paragraph{Example 2} Although vanilla BARTScore gives similar scores to these two results, we can see that VolcTrans makes an explicit error which should lead to a lower level of judgment. Specifically, VolcTrans literally translates word "\begin{CJK}{UTF8}{gbsn}多元\end{CJK}" into \textit{more yuan} (marked in \colorbox{yellow}{yellow}), which is improper because this word means \textit{diverse} in English. Fortunately, this error can be detected through error analysis, resulting in a relatively large explicit distance (0.05 compared with 0.01). Therefore, BARTScore++ can better distinguish major errors and become more human-like.

\end{document}